\def\year{2020}\relax
\documentclass[letterpaper]{article} 
\usepackage{aaai20}  
\usepackage{times}  
\usepackage{helvet} 
\usepackage{courier}  
\usepackage[hyphens]{url}  
\usepackage{graphicx} 
\usepackage{graphicx}  
\newtheorem{definition}{Definition}
\usepackage{algorithm}
\usepackage{algorithmic}
\usepackage{amsmath}
\newcommand\sysname{SPARQA}

\usepackage{color}

\title{SPARQA: Skeleton-based Semantic Parsing for Complex Questions over Knowledge Bases}

\author{Yawei Sun, Lingling Zhang, Gong Cheng, Yuzhong Qu \\
National Key Laboratory for Novel Software Technology, Nanjing University, China\\
ywsun@smail.nju.edu.cn, llzhang@smail.nju.edu.cn, gcheng@nju.edu.cn, yzqu@nju.edu.cn}

\begin{document}

\maketitle

\begin{abstract}
Semantic parsing transforms a natural language question into a formal query over a knowledge base. Many existing methods rely on syntactic parsing like dependencies. However, the accuracy of producing such expressive formalisms is not satisfying on long complex questions. In this paper, we propose a novel skeleton grammar to represent the high-level structure of a complex question. This dedicated coarse-grained formalism with a BERT-based parsing algorithm helps to improve the accuracy of the downstream fine-grained semantic parsing. Besides, to align the structure of a question with the structure of a knowledge base, our multi-strategy method combines sentence-level and word-level semantics. Our approach shows promising performance on several datasets.
\end{abstract}

\section{Introduction} \label{sect:intro}

Question answering over knowledge bases~(KBQA) has been a popular application of NLP technologies. Many recent approaches are based on semantic parsing~\cite{HerzigB18,JieL18,LabutovYM18,dong2018coarse_to_fine,dong2018confidence,chenbo2018sequence_to_action}. They transform a natural language question into a formal query, for example, a SPARQL query, which in turn is executed over a knowledge base~(KB) such as Freebase to retrieve answers. State-of-the-art methods~\cite{yanchao2018pattern_revising,mohammed2018strong,wangyue2018apva_turbo,abs-1903-02188} have achieved promising results on simple questions that are represented as a formal query with a single predicate, for example, ``who is the \emph{wife} of Obama?'' However, difficulties are faced when processing complex questions that correspond to a formal query with multiple predicates~\cite{talmor2018webcomplexq}, for example, ``what movie that Miley Cyrus \emph{acted in} had a \emph{director} named Tom Vaughan?'' We will use this question as a running example throughout the paper.

\begin{figure}
\centering
\includegraphics[width=\columnwidth]{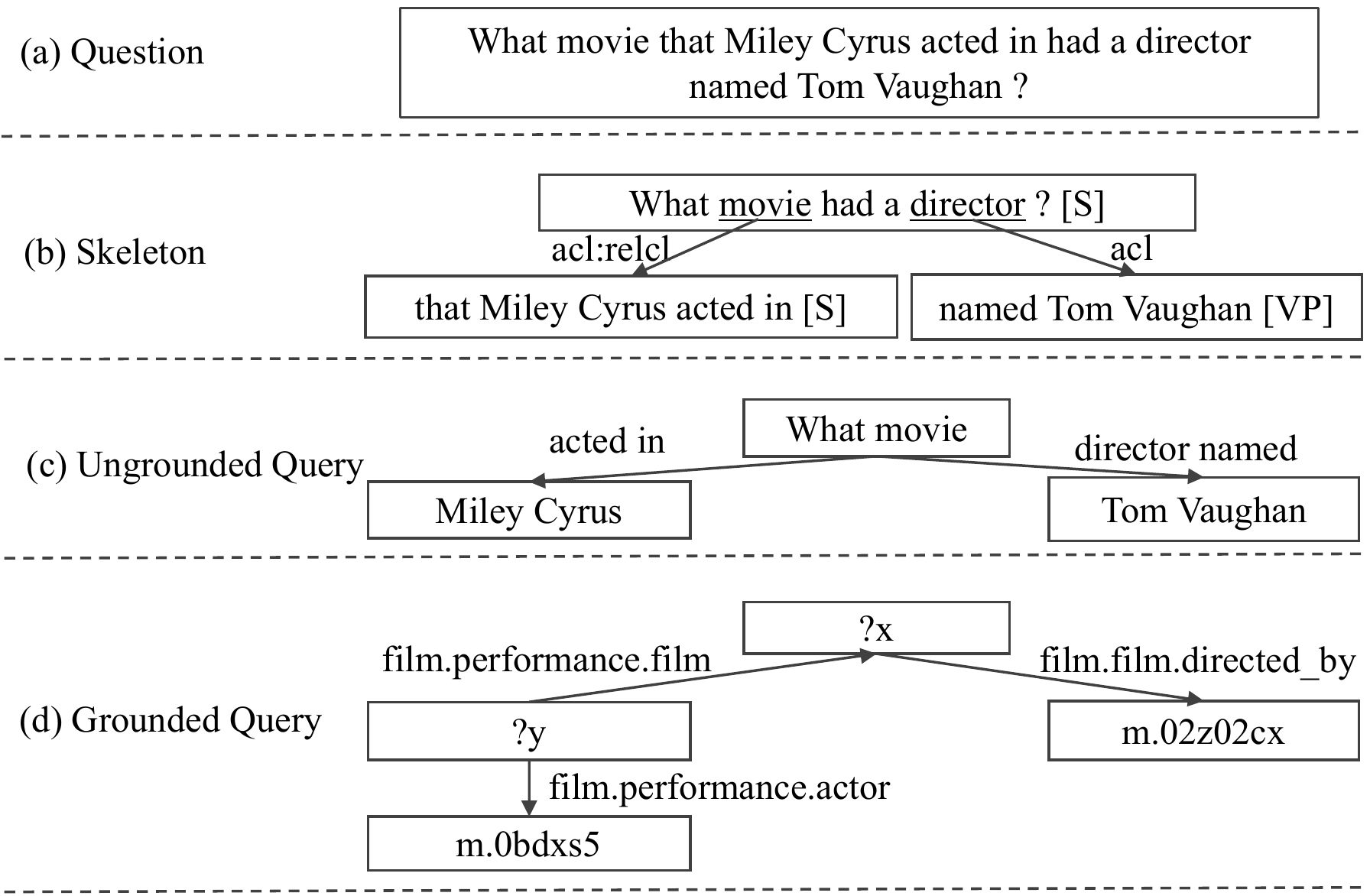}
\caption{(a)~A complex question, (b)~its skeleton, (c)~the derived ungrounded query and (d)~grounded formal query.}
\label{fig:example-skeleton}
\end{figure}

\begin{figure*}
\centering
\includegraphics[width=1.6\columnwidth]{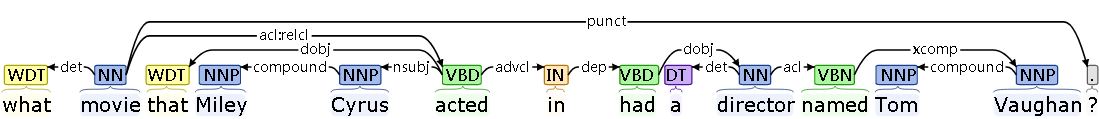}
\caption{An example of erroneous dependencies.}
\label{fig:example-motivation}
\end{figure*}

\textbf{Challenges.}
To understand and answer a complex question, we identify two challenges among others.

First, semantic parsing often relies on syntactic parsing like dependencies ~\cite{abujabal2017automated,abujabal2018never,LuoLLZ18}. Errors in syntactic parsing, which are expected for complex questions, will be propagated to the downstream semantic parsing and influence the overall performance. In our running example, as shown in Figure~\ref{fig:example-motivation}, dependency parsing misses the long-distance dependency between ``movie'' and ``had'', but generates an incorrect relation between ``in'' and ``had''.

Second, a question is often transformed into a KB-independent graph-structured ungrounded query~\cite{siva2017universal,jianpeng2017learning,husen2018subgraphmatching}, which in turn is grounded over the underlying KB into a formal query which may have a different structure. Such heterogeneity in grounding is common for complex questions with multiple predicates. In our running example, the ungrounded query in Figure~\ref{fig:example-skeleton}(c) has two predicates: \emph{acted in} and \emph{director}, but the grounded query in Figure~\ref{fig:example-skeleton}(d) has three predicates due to the use of a mediator node (\texttt{performance}) for representing n-ary relations (\texttt{actor}-\texttt{film}-\texttt{character}) in Freebase.

\begin{figure}
\centering
\includegraphics[width=0.6\columnwidth]{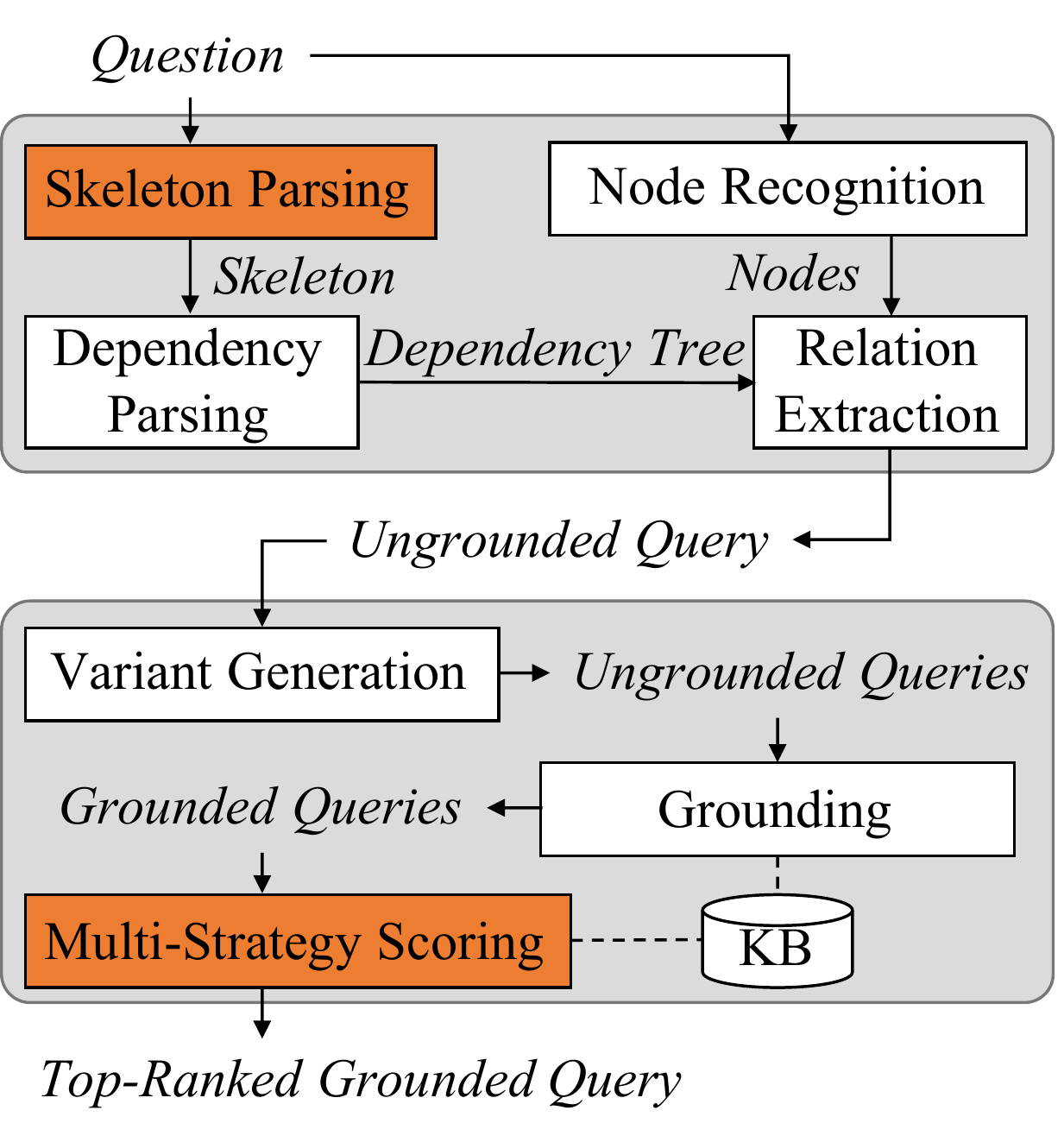}
\caption{Overview of \sysname{}.}
\label{fig:overivew}
\end{figure}

\textbf{Contributions.}
We address these two challenges with a new approach called \sysname{}, that is an abbreviation for \emph{S}keleton-based semantic \emph{PAR}sing for \emph{Q}uestion \emph{A}nswering. Figure~\ref{fig:overivew} presents an overview of our approach. From an input question, we identify its high-level skeleton structure, to help accurate generation of an ungrounded query. This ungrounded query and its structural variants are grounded against a KB. The generated grounded queries are ranked by a multi-strategy scorer, and the top-ranked grounded formal query is executed to retrieve answers from the KB. Our contribution in the paper is summarized as follows.
\begin{itemize}
    \item We propose a skeleton grammar to represent the high-level structure of a complex question. This lightweight formalism and our parsing algorithm help to improve the accuracy of the downstream semantic parsing.
    \item To train and evaluate our algorithm for skeleton parsing, we manually annotate the skeleton structure for over 10K~questions in two KBQA datasets. We make this resource public to support future research.
    \item We combine sentence-level and word-level scoring to rank grounded queries. The former mines and matches sentence patterns. The latter processes bags of words and trains a novel neural model to compute similarity.
\end{itemize}

The remainder of this paper is organized as follows. Section~\ref{sec:overview} presents an overview of \sysname{}. Section~\ref{sec:parsing} introduces skeleton parsing. Section~\ref{sec:grounding} describes multi-strategy scoring. Section~\ref{sec:experiment} reports experiment results. Section~\ref{sec:relatedwork} discusses related work. Finally, we conclude in Section~\ref{sec:conclusion}.

\section{Overview of the Approach} \label{sec:overview}

Figure~\ref{fig:overivew} presents an overview of our approach \sysname{}. Our implementation is open source.\footnote{\url{https://github.com/nju-websoft/SPARQA}}

To transform a question such as Figure~\ref{fig:example-skeleton}(a) into a KB-independent graph-structured ungrounded query in Figure~\ref{fig:example-skeleton}(c), Hu et al.~\shortcite{husen2018subgraphmatching} propose a method named NFF (short for Node-First Framework). It assumes nodes have been recognized and extracts relations between nodes based on the dependency parse of the question. Considering the standard dependency parsing of a complex question is prone to errors, we propose skeleton grammar---a subset of the dependency grammar, to represent the high-level structure of a complex question as relations between text spans, as illustrated in Figure~\ref{fig:example-skeleton}(b). The standard dependency parse of each text span are then joined to be fed into NFF for relation extraction. We will detail skeleton parsing in Section~\ref{sec:parsing}. For recognizing nodes, e.g,~mentions of entities, classes (including WH question words), and literals, we solve it as a sequence labeling problem. We use a combination of Stanford's NER~\cite{stanfordner}, SUTime~\cite{sutimeChangM12}, and a BERT-based classifier~\cite{bert}, and give priority to long mentions.

To address structural heterogeneity in grounding, we generate a set of structural variants of the original ungrounded query by contracting an edge between class nodes and/or subdividing an edge with an inserted mediator node. Each ungrounded query is grounded into a set of formal queries by linking each node to an entity, class, or literal in the KB and enumerating all possible predicates that connect adjacent nodes. For linking, we use a dictionary compiled from ClueWeb~\cite{gabrilovich2013facc1} as well as KB-specific resources. For the Freebase KB we use its topic names and aliases.
All the grounded queries are scored by a multi-strategy method, which we will detail in Section~\ref{sec:grounding}.
The top-ranked grounded formal query such as Figure~\ref{fig:example-skeleton}(d) is executed to retrieve answers from the KB.

\section{Skeleton Parsing} \label{sec:parsing}

In this section, we first introduce the skeleton grammar, and then we present our algorithm for skeleton parsing.

\subsection{Skeleton Grammar}

Our skeleton grammar is essentially a selected subset of the dependency grammar for specifically representing the high-level structure of a complex question. This dedicated coarse-grained representation, which is likely to feature an accurate parsing algorithm due to its simplicity, helps to improve the accuracy of the downstream fine-grained semantic parsing.

\begin{definition}[Skeleton]
The skeleton of a question sentence is a directed tree where nodes representing text spans in the sentence are connected by edges representing attachment relations. Specifically, a text span is attached from a headword in another span.
\end{definition}

\begin{definition}[Text Span]
Text span represents a phrase-level semantic unit. We consider four types of text spans in phrase structure grammars: Clause~(\texttt{S}), Noun Phrase~(\texttt{NP}), Verb Phrase~(\texttt{VP}), and Prepositional Phrase~(\texttt{PP}).
\end{definition}
\noindent Note that these types are only for reader comprehension. A skeleton parser is not required to label the type of a text span.

\begin{definition}[Attachment Relation]
Attachment relation represents a dependence between text spans. We consider seven types of relations that are common in standard dependency grammars:
adjectival clause~(\texttt{acl}), its sub-type relative clause modifier~(\texttt{acl:relcl}),
nominal modifier~(\texttt{nmod}), its sub-type possessive alternation~(\texttt{nmod:poss}),
coordination (\texttt{conj}),
open clausal complement~(\texttt{xcomp}), and adverbial clause modifier~(\texttt{advcl}).
\end{definition}

A skeleton has a tree structure. When we determine the granularity of our skeleton grammar and define allowed types of text spans and attachment relations, a key feature of this tree structure we want to provide is: by iteratively removing its leaf nodes, the remaining text spans in each iteration always comprise a maximal well-formed sentence, until reaching a simple sentence such that further split is not possible. This high-level structure helps to hierarchically distinguish the backbone of a complex question from other parts.

\textbf{Example.}
Figure~\ref{fig:example-skeleton}(b) shows the skeleton for our running example. The question sentence is divided into three text spans connected by two attachment relations from different headwords (underlined).

Headwords are used to join the standard dependency parse of each text span into a full dependency tree for the original question, to be fed into the downstream semantic parsing as described in Section~\ref{sec:overview}. Such a two-stage dependency parsing is expected to feature improved accuracy as the skeleton parsing in the first stage adopts a lightweight formalism, and the standard dependency parsing in the second stage processes simple text spans.

\begin{algorithm}[!t]
\caption{Skeleton Parsing}
\label{alg:sp}
\begin{algorithmic}
\REQUIRE A sentence~$Q$
\ENSURE The skeleton of~$Q$
\STATE $T \leftarrow$ tree with a root node~$Q$
\WHILE{Split($Q$) is true}
  \STATE $s \leftarrow$ TextSpanPrediction($Q$)
  \STATE $h \leftarrow$ HeadwordIdentification($s$, $Q$)
  \STATE $r \leftarrow$ AttachmentRelationClassification($s$, $Q$)
  \STATE Remove~$s$ from~$Q$
  \STATE Grow~$T$ with relation~$r$ from~$h \in Q$ to~$s$
\ENDWHILE
\RETURN $T$
\end{algorithmic}
\end{algorithm}

\subsection{Algorithm for Skeleton Parsing}

Algorithm~\ref{alg:sp} parses a question sentence~$Q$ into its skeleton denoted by~$T$. Initially, $T$~comprises a single root node representing the entire sentence~$Q$ as a text span. Then iteratively, we grow~$T$ by splitting~$Q$ and adding a new edge. The Split procedure decides whether $Q$~needs further split. If needed, the TextSpanPrediction procedure predicts the next text span to be split from~$Q$, denoted by~$s$. It will be removed from~$Q$ and attached from a headword~$h$ in the remaining~$Q$, which is identified by the HeadwordIdentification procedure. The AttachmentRelationClassification procedure determines the attachment relation. After iterations, $T$~is returned when $Q$~needs no further split.

\textbf{Example.}
In Figure~\ref{fig:example-skeleton}, text span ``named Tom Vaughan'' is firstly split from the question and attached from its headword ``director'' in the remaining question with relation~\texttt{acl}. Then, text span ``that Miley Cyrus acted in'' is split and attached from ``movie'' with relation~\texttt{acl:relcl}. The remaining ``what movie had a director'' needs no further split, so skeleton parsing is completed. Note that a skeleton is generally not limited to have a star structure as in this example.

The four procedures mentioned above are implemented based on BERT~\cite{bert}, where the authors illustrated fine-tuning BERT on several tasks:
\begin{itemize}
    \item SPC: sentence pair classification,
    \item SSC: single sentence classification, and
    \item QA: question answering.
\end{itemize}

\begin{figure*}
\centering
\includegraphics[width=1.75\columnwidth]{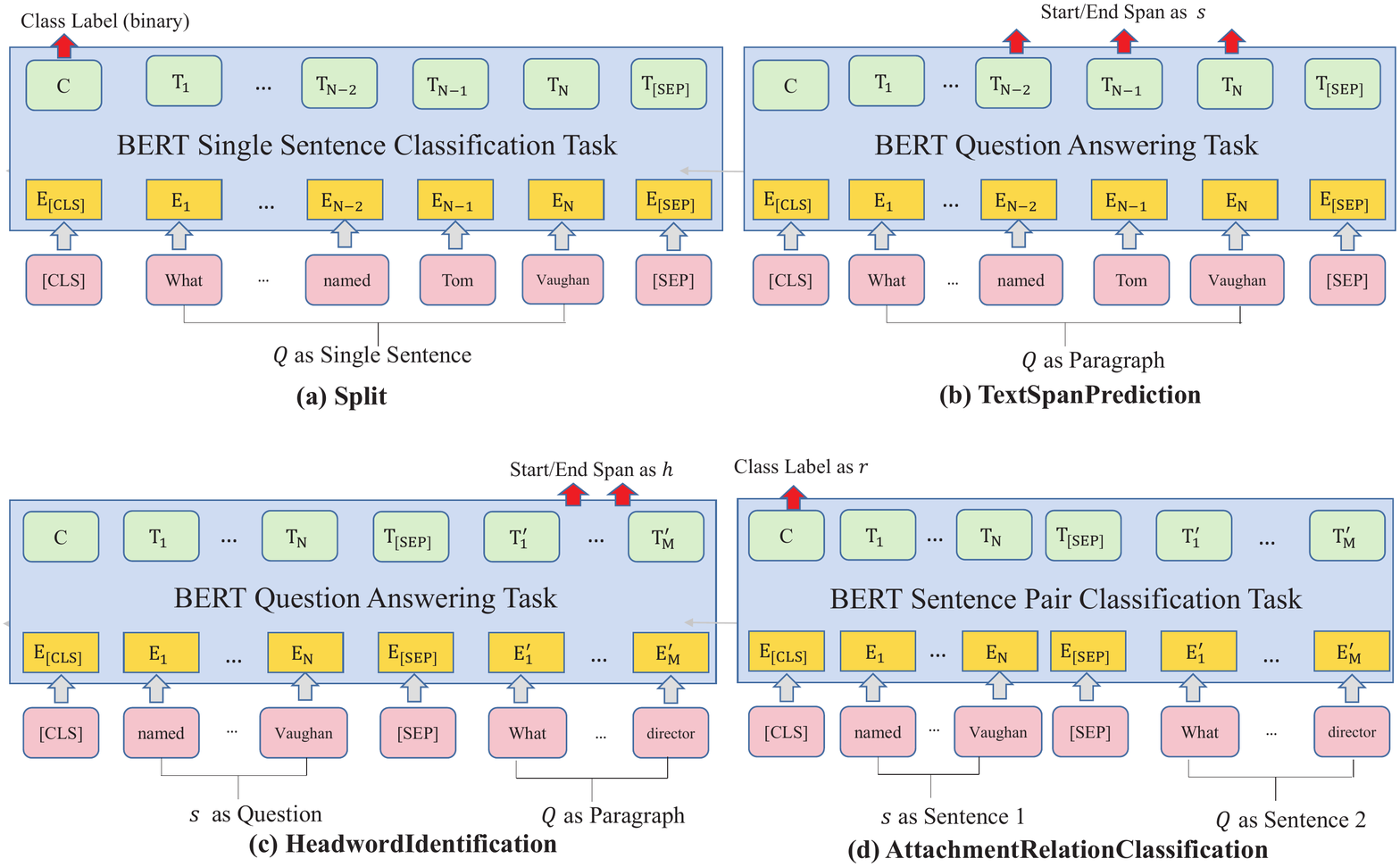}
\caption{BERT-based implementation of skeleton parsing.}
\label{fig:bert}
\end{figure*}

\noindent Our BERT-based implementation of the four procedures is shown in Figure~\ref{fig:bert} and detailed below. For our experiments, we manually annotated the skeleton structure for questions in the training set to fine-tune our BERT models.

\textbf{Split.}
This procedure decides whether $Q$~needs further split. We formulate it as an SSC task that is supported by a standard fine-tuned BERT model. We treat~$Q$ as the single sentence fed into the model, which outputs a binary value.

\textbf{TextSpanPrediction.}
This procedure predicts the next text span~$s$ to be split from~$Q$. We formulate it as a QA task that is supported by a standard fine-tuned BERT model. We disable the question input and treat~$Q$ as the paragraph fed into the model, which outputs a span in~$Q$ as~$s$.

\textbf{HeadwordIdentification.}
This procedure identifies a headword in the remaining~$Q$ from which $s$~is attached. We formulate it as a QA task that is supported by a standard fine-tuned BERT model. We treat the remaining~$Q$ as the paragraph and treat~$s$ as the question fed into the model, which outputs a span (restricted to a single word) in the remaining~$Q$ as~$h$.

\textbf{AttachmentRelationClassification.}
This procedure determines the attachment relation~$r$ from~$h$ in the remaining~$Q$ to~$s$. We formulate it as an SPC task that is supported by a standard fine-tuned BERT model. We treat~$s$ and the remaining~$Q$ as two sentences fed into the model, which outputs one of the seven predefined attachment relations as~$r$.

\section{Multi-Strategy Scoring} \label{sec:grounding}

Candidate grounded queries are ranked by their total scores output by a sentence-level scorer and a word-level scorer, which we describe in this section.

\subsection{Sentence-Level Scorer}

This scorer exploits known mappings from questions to formal queries by mining and matching sentence/query patterns. The pattern of a question is obtained by replacing entity mentions with dummy tokens. For a question in the training set, the pattern of its underlying formal query is obtained by replacing entities that correspond to dummy tokens in the question with placeholders.

\textbf{Example.}
The pattern of our running example is ``what movie that $\left \langle E_1 \right \rangle$ acted in had a director named $\left \langle E_2 \right \rangle$?'' If this is a question in the training set, the pattern of its underlying formal query is derived from Figure~\ref{fig:example-skeleton}(d) by replacing ``m.0bdxs5'' and ``m.02z02cx'' with placeholders.

Given a test question, we find its most similar question in the training set that has the same number of dummy tokens in their patterns. We replace the placeholders in the pattern of this training question's underlying formal query with the corresponding entities in the test question, to generate a grounded formal query. This query will score~1.0 if it retrieves non-empty results, or~0.0 otherwise. All the other candidate grounded queries trivially score~0.0.

Similarity is computed by a standard fine-tuned BERT model that supports the SPC task, which is fed with the patterns of two sentences and outputs a numerical value representing similarity. During training, we use pairs of questions in the training set that have the same question pattern and the same query pattern as positive examples, and use other random pairs as negative examples. During testing, we use the fine-tuned model to predict the similarity between the test question and each training question.

\begin{figure}
\centering
\includegraphics[width=0.8\columnwidth]{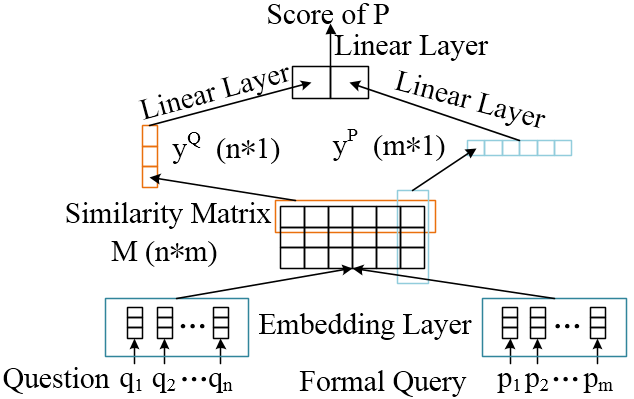}
\caption{Word-level scorer.}
\label{fig:path_match}
\end{figure}

\subsection{Word-Level Scorer}

This scorer is based on bag of words. We train a novel neural model shown in Figure~\ref{fig:path_match} to score a grounded formal query. Specifically, we represent questions and formal queries as bags of words, where we remove concrete entities and stop words. The remaining words mainly represent predicates. Let $Q=\{q_1,\ldots,q_n\}$ be a bag of words for a question. Let $P=\{p_1,\ldots,p_m\}$ be a bag of words for a formal query. In our model, words are converted into their pre-trained GloVe embeddings~\cite{pennington2014glove}. We calculate the cosine similarity between pairs of embedding vectors in~$Q$ and~$P$, forming an $n \times m$ similarity matrix denoted by $M=[M_{i,j}]_{n*m}$. We take the maximum value in each row and in each column, producing an $n$-dimensional vector and an $m$-dimensional vector, denoted by~$\mathbf{y}^Q$ and~$\mathbf{y}^P$, respectively:
\begin{equation*}
\begin{split}
    \mathbf{y}^Q & = [y^Q_i]_n \,, \quad y^Q_i = \max\limits_{1 \le j \le m} {M_{i,j}} \,, \\
    \mathbf{y}^P & = [y^P_j]_m \,, \quad y^P_j = \max\limits_{1 \le i \le n} {M_{i,j}} \,.
\end{split}
\end{equation*}
\noindent We feed~$\mathbf{y}^Q$ and~$\mathbf{y}^P$ into linear layers, which in turn are fed into another linear layer to output a numerical score for the formal query.

During training, we use questions in the training set and their underlying formal queries as positive examples, and use other random combinations as negative examples. During testing, we use the learned model to predict the score of each candidate grounded formal query.

\section{Experiments} \label{sec:experiment}
In this section, we evaluate our approach \sysname{} on two standard KBQA datasets with complex questions. We compare with baselines, perform an ablation study of our approach, and finally we analyze each component of our skeleton parser. Note that as our implementation relies on an existing method for fine-grained semantic parsing~\cite{husen2018subgraphmatching}, our experiments will give more attention to the effectiveness of our proposed add-on (i.e.,~skeleton parsing) rather than the overall performance of our full approach.

\subsection{Datasets}\label{ssec:dataset}
The experiments were performed on two public datasets involving complex questions.
\begin{itemize}

    \item \textbf{GraphQuestions}~\cite{su2016graphquestion} contains 5,166~questions---2,558~for training and 2,608~for testing.
    They can be transformed into SPARQL queries over Freebase (version June 2013).

    \item \textbf{ComplexWebQuestions} version~1.1~\cite{talmor2018webcomplexq} contains 34,689 questions with a split of 80-10-10 for training, validation, and test sets.
    They can be transformed into SPARQL queries over Freebase (version 2015-08-09). Alternatively, they can also be answered based on provided search engine snippets.

\end{itemize}

\subsection{Baselines}\label{ssec:baselines}

On GraphQuestions, we compared with six methods with reported results on this dataset.
\begin{itemize}
  \item \textbf{SEMPRE}~\cite{berant2013semantic} is a bottom-up beam-based semantic parsing method.
  \item \textbf{PARASEMPRE}~\cite{berant2014semantic} is a paraphrase-based semantic parsing method.
  \item \textbf{JACANA}~\cite{YaoD14} is an information extraction based method.
  \item \textbf{UDEPLAMBDA}~\cite{siva2017universal} is a dependency-based semantic parsing method.
  \item \textbf{SCANNER}~\cite{jianpeng2017learning} is a transition-based neural semantic parsing method.
  \item \textbf{PARA4QA}~\cite{dong2017paraphrase} is a paraphrase-based neural method.
\end{itemize}

On ComplexWebQuestions, we compared with five methods with reported results on this dataset. Note that this dataset also provided relevant search engine snippets.
\begin{itemize}
    \item \textbf{MHQA-GRN}~\cite{cwq-yuezhang} uses graph neural networks to answer a question as a reading comprehension task over snippets.
    \item \textbf{SIMPQA + PRETRAINED}~\cite{talmor2018webcomplexq} uses a pre-trained reading comprehension model over snippets.
    \item \textbf{SPLITQA + PRETRAINED}~\cite{talmor2018repartitioning} uses the same pre-trained model as SIMPQA+PRETRAINED but it decomposes a question and then recomposes the final answer.
    \item \textbf{SPLITQA + data augmentation}~\cite{talmor2018repartitioning} uses the same method as SPLITQA + PRETRAINED but is trained on ComplexWebQuestions examples as well as an additional large set of examples that are not accessible to our approach.
    \item \textbf{PullNet}~\cite{pullnet} uses graph convolutional networks to extract answers from KB and/or snippets.
\end{itemize}

\subsection{Configuration of \sysname{}}\label{ssec:detail}

\textbf{Skeleton Parser.}
To train and test our skeleton parser, we manually annotated the skeleton structure for all the 5,166~questions in GraphQuestions and 5,000~questions in ComplexWebQuestions: 3,000~from the training set, 1,000~from the validation set, and 1,000~from the test set. We have made this resource public to support future research.\footnote{\url{https://github.com/nju-websoft/SPARQA}}

In our skeleton parser, all the four BERT models were based on $\text{BERT}_\text{BASE}$~(L =~12, H =~768, A =~12, total parameters =~110M). Their hyperparameters were:
\begin{itemize}
	\item Split: max sequence length =~32, learning rate =~3e-5, batch size =~32, training epochs =~100,
	\item TextSpanPrediction: max sequence length =~32, learning rate =~3e-5, batch size =~32, training epochs =~100, 
	\item HeadwordIdentification: max sequence length =~32, learning rate =~3e-5, batch size =~32, training epochs =~100,
	\item AttachmentRelationClassification: max sequence length =~64, learning rate =~4e-5, batch size =~32, training epochs =~100. 
\end{itemize}

\textbf{Sentence-Level Scorer.}
The BERT model was configured as follows:
\begin{itemize}
	\item max sequence length =~64, learning rate =~3e-5, batch size =~32, training epochs =~4.
\end{itemize}

\textbf{Word-Level Scorer.}
We used 300-dimensional pre-trained GloVe embeddings. The neural model was trained with hinge loss with negative sampling size =~300, using Adam with learning rate =~0.001
and batch size=~32.

\subsection{Overall Results} \label{ssec:results}
Following common practice in the literature, we report F1 on GraphQuestions and report Precision@1 (P@1) on ComplexWebQuestions averaged over all the test questions.

\begin{table}[t!]
\begin{center}
\begin{tabular}{|lr|}
\hline
& \bf F1 \\ 
\hline
SEMPRE & 10.80 \\
PARASEMPRE & 12.79 \\
JACANA & 5.08 \\
UDEPLAMBDA & 17.70 \\
SCANNER & 17.02 \\
PARA4QA & 20.40 \\
\hline
SPARQA & 21.53 \\ 
\hline
\end{tabular}
\end{center}
\caption{\label{tab:graphq-end-to-end} Overall results on GraphQuestions.}
\end{table}

The results on GraphQuestions are summarized in Table~\ref{tab:graphq-end-to-end}. Around half of the questions in this dataset are complex questions. Others are simple questions. Our \sysname{} achieved a new state-of-the-art result on this dataset, outperforming all the known baseline results. F1 was improved from~20.40---the previous best result achieved by the PARA4QA, to~21.53, by an increase of~5.5\%. We would like to highlight the result~17.70 achieved by UDEPLAMBDA, which also adopted dependency-based semantic parsing but relied on a set of predefined rules. It demonstrated the effectiveness of our skeleton-based semantic parsing. Moreover, our skeleton parsing did not hurt accuracy on the 1,172~simple questions in this dataset, where \sysname{} (F1=27.68) was comparable with PARA4QA (F1=27.42).

\begin{table}[t!]
\begin{center}
\begin{tabular}{|lr|}
\hline
& \bf P@1 \\
\hline
MHQA-GRN & 30.10 \\
SIMPQA + PRETRAINED & 19.90 \\
SPLITQA + PRETRAINED & 25.90 \\
SPLITQA + data augmentation & 34.20 \\
PullNet & 45.90 \\
\hline
SPARQA & 31.57 \\
\hline
\end{tabular}
\end{center}
\caption{\label{tab:cwq-end-to-end} Overall results on ComplexWebQuestions.}
\end{table}

The results on ComplexWebQuestions are summarized in Table~\ref{tab:cwq-end-to-end}. All the questions in this dataset are complex questions. Our \sysname{} outperformed most baselines except for two that used additional data. Specifically, SPLITQA + data augmentation used additional 28,674~magic training examples that were not accessible to our approach. PullNet used not only KB but also external search engine snippets which were not used in our approach.

\subsection{Ablation Study}\label{ssec:ablation}
We analyzed the usefulness of our two technical contributions: skeleton parsing and multi-strategy scoring.

\textbf{Skeleton Parsing.}
Recall that in \sysname{}, skeleton parsing is performed to generate more accurate dependencies, which are then fed into an existing method~\cite{husen2018subgraphmatching} to generate an ungrounded query. To explore the usefulness of skeleton parsing, we removed skeleton parsing and directly fed the dependencies generated by Stanford CoreNLP into the downstream module.

The results on ComplexWebQuestions are shown in Table~\ref{tab:woskeleton-cwq}. By excluding skeleton parsing, P@1 decreased notably from~31.57 to~29.39 by~6.9\%. As all the questions in this dataset are complex questions, the result demonstrated the usefulness of our proposed skeleton parsing in improving the accuracy of dependency-based semantic parsing.

\textbf{Multi-Strategy Scoring.}
In \sysname{}, two methods are implemented to score candidate formal queries: a sentence-level scorer and a word-level scorer. Their scores are combined in the end. To explore their usefulness, we removed either of them and only used the other.

The results on ComplexWebQuestions are shown in Table~\ref{tab:woskeleton-cwq}. By excluding the sentence-level scorer, P@1 decreased considerably from~31.57 to~26.45 by~16.2\%. By excluding the word-level scorer, the decrease was even larger, from~31.57 to~26.11 by~17.3\%. The results demonstrated the usefulness of our proposed two scorers in improving the accuracy of semantic parsing.

\begin{table}[t!]
\begin{center}
\begin{tabular}{|lr|}
\hline
& \bf P@1 \\ 
\hline
\sysname{} & 31.57 \\ 
\hline
\sysname{} w/o skeleton parsing & 29.39 \\ 
\sysname{} w/o sentence-level scorer & 26.45 \\ 
\sysname{} w/o word-level scorer & 26.11 \\ 
\hline
\end{tabular}
\end{center}
\caption{\label{tab:woskeleton-cwq} Ablation study of \sysname{}.}
\end{table}

\begin{table}[t!]
\begin{center}
\begin{tabular}{|lr|}
\hline
Overall LAS  & 93.73 \\
\hline
Accuracy of Split & 99.42\\
Accuracy of TextSpanPrediction & 97.17 \\
Accuracy of HeadwordIdentification & 97.22\\
Accuracy of AttachmentRelationClassification & 99.14\\
\hline
\end{tabular}
\end{center}
\caption{\label{tab:procedures-cwq} Accuracy of skeleton parsing.}
\end{table}

\subsection{Accuracy of Skeleton Parsing}

The above ablation study has demonstrated the usefulness of our skeleton parsing in an extrinsic manner---through the KBQA task. Below we show the results of intrinsic evaluation by comparing the output of our skeleton parser with manually annotated gold standard for the 1,000~test questions in ComplexWebQuestions.

The results are presented in Table~\ref{tab:procedures-cwq}. We computed LAS---Labeled Attachment Score, a commonly used metric for evaluating dependency parsing. The overall LAS of our skeleton parsing reached~93.73. We also evaluated the accuracy of each of the four components of our skeleton parser. The results were satisfyingly all above~97\%. The results demonstrated the effectiveness of our BERT-based parsing algorithm. With this dedicated skeleton structure for representing complex questions and our accurate parsing algorithm, the performance of the downstream semantic parsing was therefore improved in the ablation study (Table~\ref{tab:woskeleton-cwq}).

\subsection{Error Analysis} \label{ssec:error}

We randomly sampled 100~questions where our \sysname{} achieved P@1=0 on ComplexWebQuestions. We classified the errors into the following categories.

\emph{Node Recognition and Linking} (37\%): It was mainly related to long mentions that were hard to recognize, for example, the entity mention ``Rihanna: Live in Concert Tour'' in the question ``where was the artist had a concert tour named Rihanna: Live in Concert Tour raised?''

\emph{Skeleton Parsing} (5\%): One typical error was in headword identification. Long-distance attachment was sometimes not found. For example, the text span ``with a capital called Brussels'' in the question ``What country speaks Germanic languages with a capital called Brussels?'' should be attached from headword ``country'', but was mistakenly linked to ``languages'' by our parser.

\emph{Ungrounded Query} (10\%): It was caused by the off-the-shelf method we used to generate ungrounded queries from dependencies~\cite{husen2018subgraphmatching}.

\emph{Structural Heterogeneity} (22\%): For example, for the question ``who is the prime minister of the country that has national anthem March Forward, Dear Mother Ethiopia'', its skeleton and the derived ungrounded query were path-structured, but the correct formal query over Freebase was actually star-structured.

\emph{Scoring} (15\%):
Our sentence-level scorer failed when the training set did not cover a test question's query pattern. Another case was more challenging. The test question and the training question had the same pattern, but corresponded to formal queries of different patterns. Our word-level scorer failed when the pattern of a question contained very few content words, for example, ``what about $\left \langle E \right \rangle$?''. Other errors were related to polysemy and synonyms.

\emph{Others} (11\%):
Other errors were related to aggregate questions which we did not specifically process, as well as typos, character encoding issues, etc.

\section{Related Work} \label{sec:relatedwork}

\textbf{Semantic Parsing} has gained increasing research attention. Traditional rule-based methods are limited by their generalizability~\cite{LiangJK11,berant2013semantic,berant2014semantic}. Transition-based stated parsing defines states and actions to search for possible formal queries~\cite{yih2015semantic,jianpeng2017learning,Hu0Z18,chenbo2018sequence_to_action}. Recent neural methods employ encoder-decoder models to solve semantic parsing as a sequence transduction problem~\cite{JiaL16,GuptaL18,sorokin2018modeling,dong2018coarse_to_fine,dong2018confidence,LuoLLZ18}. By comparison, dependency-based methods transform the dependency structure of a question into a formal query, thus being more explainable~\cite{siva2016transforming,siva2017universal,abujabal2017automated,husen2018subgraphmatching}. However, their performance is related to the performance of dependency parsing, which is not satisfying on complex questions. In our \sysname{}, more accurate dependencies are obtained by joining local dependencies according to a global skeleton structure. The higher accuracy is attributed to the lightweight formalism of the skeleton grammar and the simplicity of the text spans where dependency parsing is performed.

\textbf{Predicate Mapping} is a key step in question answering over knowledge bases. Early feature-rich methods exploit lexicons and syntactic information~\cite{yao2014information,bast2015more}. Recent neural methods encode questions and predicates for computing their similarity. They encode predicates at different granularities, for example, at the character level~\cite{yih2015semantic,yin2016simple} or at the word level~\cite{yu2017improved,LuoLLZ18}. Various models have been used, from the simple CNN~\cite{yih2015semantic,yin2016simple} to more complex models that combine multi-level representations and similarity~\cite{yu2017improved}. By comparison, we focus on sentence-level and word-level semantics. Our novel implementation shows effectiveness in the ablation study.

\section{Conclusion and Future Work} \label{sec:conclusion}

Our skeleton parsing has shown usefulness in processing complex questions. With our proposed skeleton structure of a question, more accurate dependencies are derived, which in turn benefit the downstream fine-grained semantic parsing. Our skeleton parser can be combined with other dependency-based KBQA methods, not limited to the one used in our implementation. It may also find application in other tasks where complex sentences are common. Besides, our simple yet effective word-level scoring model can also be used as a generic similarity measure. We will experiment with these extensions in future work.

Experiments reveal the following limitations of our work to be overcome in future work. First, node recognition and linking, though being out of the scope of our contribution, is a major weakness of our full approach. It is still an open problem in question answering research. Second, for structural heterogeneity, one may explore some graph-based methods, such as graph edit distance and graph neural network. Third, aggregate questions are not the focus of our approach, but occupy a considerable proportion in complex questions. It may be possible to learn templates for parsing such questions.

\section{Acknowledgments}
This work was supported in part by the National Key R\&D Program of China under Grant 2018YFB1005100, and in part by the NSFC under Grant 61772264. Gong Cheng was supported by the Qing Lan Program of Jiangsu Province.

\bibliography{AAAI-SunY.3419}
\bibliographystyle{aaai}

\end{document}